%% file: ISALux-main.tex
\documentclass{article}

\usepackage{arxiv}
\usepackage{graphicx}
\usepackage{booktabs}
\usepackage[dvipsnames]{xcolor}
\usepackage{amsmath,amssymb,amsfonts}
\usepackage{url}
\usepackage{xcolor}
\usepackage{multirow} 
\usepackage{multicol} 
\usepackage{makecell} 
\usepackage{enumitem}
\usepackage{colortbl}
\usepackage{nicematrix}
\usepackage{caption}
\usepackage{subcaption}

\title{ISALux: Illumination and Semantics-Aware Transformer Employing Mixture of Experts for Low Light Image Enhancement}

\author{
Raul Balmez \\
  Department of Computer Science \\
  University of Manchester \\
  Manchester, UK \\
  \texttt{raul.balmez@student.manchester.ac.uk} \\
  \And
Alexandru Brateanu \\
  Department of Computer Science \\
  University of Manchester \\
  Manchester, UK \\
  \texttt{alexandru.brateanu@student.manchester.ac.uk} \\
  \And
Ciprian Orhei \\
  Department of Computer and Information Technology \\
  Politehnica University of Timisoara \\
  Timisoara, Romania \\
  \texttt{ciprian.orhei@upt.ro} \\
  \And
Codruta Ancuti \\
  Department of Computer and Information Technology \\
  Politehnica University of Timisoara \\
  Timisoara, Romania \\
  \texttt{codruta.ancuti@upt.ro} \\
  \And
Cosmin Ancuti \\
  Department of Computer and Information Technology \\
  Politehnica University of Timisoara \\
  Timisoara, Romania \\
  \texttt{cosmin.ancuti@upt.ro} \\
}

\begin{document}

\maketitle

\input{sec/0_abstract} 
\input{sec/1_intro}
\input{sec/2_related}
\input{sec/3_methods}

\input{sec/4_results}
\input{sec/5_ablation}

\input{sec/6_conclusion}
\bibliographystyle{unsrt}

\bibliography{egbib}
\end{document}

%% file: sec/0_abstract.tex
\begin{abstract}
We introduce ISALux, a novel transformer-based approach for Low-Light Image Enhancement (LLIE) that seamlessly integrates illumination and semantic priors. Our architecture includes an original self-attention block, Hybrid Illumination and Semantics-Aware Multi-Headed Self-Attention (HISA-MSA), which integrates illumination and semantic segmentation maps for enhanced feature extraction. ISALux employs two self-attention modules to independently process illumination and semantic features, selectively enriching each other to regulate luminance and highlight structural variations in real-world scenarios. A Mixture of Experts (MoE)-based Feed-Forward Network (FFN) enhances contextual learning, with a gating mechanism conditionally activating the top K experts for specialized processing. To address overfitting in LLIE methods caused by distinct light patterns in benchmarking datasets, we enhance the HISA-MSA module with low-rank matrix adaptations (LoRA). Extensive qualitative and quantitative evaluations across multiple specialized datasets demonstrate that ISALux is competitive with state-of-the-art (SOTA) methods. Additionally, an ablation study highlights the contribution of each component in the proposed model. Code will be released upon publication.
\end{abstract}

%% file: sec/1_intro.tex
\section{Introduction}
\label{sec:intro}


\noindent Illumination is a crucial factor in images, affecting color representation, local and global structures, and overall visibility of visual information. Low-light image enhancement (LLIE) is vital for restoration and extracting important features under uneven lighting~\cite{HE_review}. However, low-light introduces challenges—noise, color distortion, misinterpretation, artifacts—and undermines computer vision applications like object detection that rely on high-quality input. Without clear, informative representations, these tasks struggle to achieve accurate results.



The rise of deep learning (DL) has greatly advanced low-light image enhancement (LLIE)~\cite{LLIE_survey_2022}. CNN-based models~\cite{DeepUPE,DeepLPF,Uformer,RetinexNet} initially led this progress, laying the foundation for more advanced techniques. Recently, Transformers have emerged as strong alternatives, redefining image restoration. Unlike CNNs, Vision Transformers (ViTs)~\cite{Transformer_NIPS2017} use self-attention to capture long-range dependencies, offering advantages in LLIE. However, current methods often apply uniform enhancements across all pixels, overlooking structural image cues. As emphasized by Li et al.~\cite{LLIE_survey_2022}, integrating semantic information is vital. In particular, semantic segmentation enables region-aware enhancement, addressing local illumination variations and improving overall quality.



\noindent To address this, we introduce ISALux, a method for real-world scenarios where different objects require varying illumination levels. We design a novel self-attention block, Hybrid Illumination and Semantics-Aware Multi-Headed Self-Attention (HISA-MSA), integrating illumination and semantic information. ISALux uses two self-attention modules for independent extraction: illumination and semantics. Each attention map is selectively enriched with the other—illumination for global lighting context, semantics for structural variations. To mitigate overfitting from diverse light patterns, we enhance HISA-MSA with low-rank adaptations (LoRA) \cite{hu2022lora}, employed as regular trainable parameters to improve enhancement under varying illumination during self-attention.

To enhance contextual feature extraction within the HISA-MSA, the Feed-Forward Network (FFN) employs a Mixture of Experts (MoE) architecture. This design enables the FFN to adaptively learn to address different deficiencies introduced by the absence of light. 
A gating mechanism assigns selection probabilities to each expert, enabling the model to activate only the top K experts.
Each expert operates as an independent CNN, facilitating localized feature refinement and specialized processing. 
The proposed architecture ensures efficient adaptability to diverse and challenging contexts.

Our extensive qualitative and quantitative evaluation across multiple specialized datasets demonstrates that ISALux is competitive with state-of-the-art (SOTA). Finally, we present an ablation study to evaluate the contribution of each component within the proposed model.


Contributions of ISALux can be summarized as follows:

\begin{itemize}
\item ISALux presents a novel way to balance the enhancement process for LLIE using a ViT-based encoder-decoder that fuses semantic and global illumination cues.

\item We propose HISA-MSA which enriches self-attended maps by integrating illumination and segmentation priors.

\item Moreover, the MoE architecture is leveraged to enhance the performance of HISA-MSA while maintaining computational efficiency.

\item ISALux achieves SOTA results on key benchmarks, proving its robustness on both reference and non-reference datasets.
\end{itemize}

%% file: sec/2_related.tex
\section{Related work}
\label{sec:relateed}



\textbf{CNN-based methods.} With the rapid progress of deep learning, convolutional neural networks (CNNs) have been extensively employed for low-light image enhancement~\cite{RetinexNet, DeepLPF, DeepUPE, KinD, Zhang_2020, Diff-retinex}. Retinex-Net~\cite{RetinexNet} was among the first deep networks to incorporate both image decomposition and enhancement for this task. However, CNN-based approaches often struggle to effectively capture long-range dependencies, limiting their ability to model global image structures.



\textbf{ViTs techniques}. To address the limitations of convolutional neural networks (CNNs), Vision Transformers (ViTs)~\cite{Dosovitskiy_2021} have been successfully employed in a wide range of computer vision applications ~\cite{Yuan_ICCV_2021, Wang_ICCV_2021, Zheng_CVPR_2021, Cariong_ECCV_2020, Liu_ICCV_2021}. Furthermore, the self-attention (SA) mechanism in ViTs has been extensively explored for various image restoration tasks~\cite{Yang_CVPR2020,X_Chen_CVPR2023,Dehazing_Trans,yang2024dvt} , including LLIE~\cite{SNR-Net,Zhang_ICCV2021,Cui_BMVC2022,Retinexformer}. The recently proposed Retinexformer~\cite{Retinexformer}   leverages illumination representations to facilitate the modeling of non-local interactions across regions with varying lighting conditions. Similarly, ISALux is a transformer-based architecture, however, it modifies the SA block to integrate both illumination and semantic segmentation maps, enhancing its capacity for LLIE task.


\textbf{Mixture of Experts (MoE). } is a powerful and adaptable framework designed to improve the performance and computational efficiency of DL models. Recent rise of MoE in popularity has been largely driven by the advent of Large Language Models (LLMs) and the increasing prevalence of transformer-based architectures in modern machine learning research. MoE has been widely used in LLMs~\cite{GLAM_2022, Raphael_2023, li2024cumo, kang2024selfmoecompositionallargelanguage} and high-level vision tasks~\cite{Chen_NIPS2022, Chowdhury_2023, AdaMV-MoE, ACE_2021}. However, only a few approaches have been explored for image restoration ~\cite{Wang_2020, MoESR_2022, guoparameter_2024}. In contrast, our architecture leverages MoE specifically for LLIE, where all input pixels are processed by experts selected through a gating mechanism.


%% file: sec/3_methods.tex
\section{Our Approach}

\begin{figure*}[ht!]
        \centering
        \includegraphics[trim=0cm 0cm 0cm 0cm, clip=true, width=0.96\linewidth]{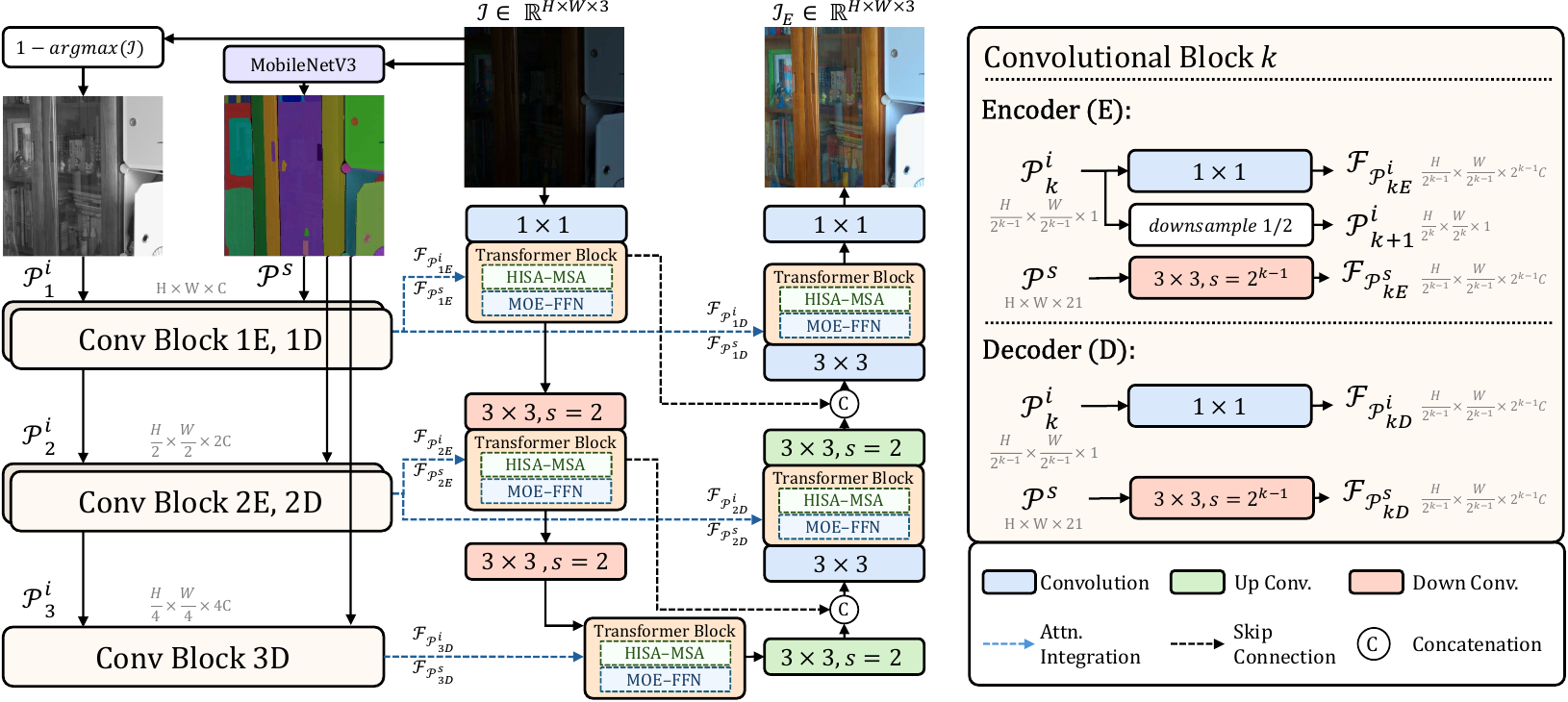}

        \caption{A visual representation of the \textit{ISALux}'s architecture. The method follows an encoder-decoder (ViT) framework. Additionally, we illustrate the computation and preprocessing of illumination and semantic priors, which are subsequently incorporated into each Transformer Block to enrich the attention map's representations.}
        \label{fig:framework}
        
    \end{figure*}

\subsection{The illumination prior}

Our method is based on the use of priors to compensate for the lack of fine-details and structure introduced by the light-deficient images. The illumination component integrated in the model architecture, successfully used previously by \cite{Retinexformer}, has the role of conferring a local and global context of the light distribution, which is used in the HISA-MSA. Therefore, starting from an input image $\mathcal{I} \in \mathbb{R}^{H\times W \times3}$, we assume the illumination prior to be:

\begin{equation}
\mathcal{P}^{i} = 1 - \arg\max\limits_{c}(\mathcal{I}), \quad \mathcal{P}^{i} \in \mathbb{R}^{H \times W \times 1}
\end{equation}

\noindent where $\arg\max_{c}$ returns the maximum channel value for each input pixel. The following stage is the computation of the illumination prior pyramid starting from $\mathcal{P}^{i}$, which comprises the initial prior and two downsampled features: $\mathcal{P}^{i}_{0}$, $\mathcal{P}^{i}_{1}$, $\mathcal{P}^{i}_{2}$, where:

\begin{gather}
\mathcal{P}^{i}_{0} = \mathcal{P}^{i}; \quad \mathcal{P}^{i}_{1} = \mathcal{F}_{\downarrow 2}(\mathcal{P}^{i}); \quad \mathcal{P}^{i}_{2} = \mathcal{F}_{\downarrow 4}(\mathcal{P}^{i})
\end{gather}

\subsection{The semantic prior}
The semantic prior provides spatial information and enriches the representations of the self-attended features. As the semantic segmentation backbone of ISALux, we use MobileNetV3~\cite{MobilNetV3} for its robustness and computational efficiency, comprising only $11.03$M parameters. The model is trained on a subset of the COCO dataset~\cite{lin2014microsoft}. Starting from the input image, $\mathcal{I} \in \mathbb{R}^{H\times W \times3}$, we consider the semantic prior to be $\mathcal{P}^{s} \in  \mathbb{R}^{H \times W \times 21}$, where each of the $21$ channels corresponds to the probability map of a specific class.

\begin{equation}
\mathcal{P}^{s} = \textbf{MobileNetV3}(\mathcal{I}), \quad \mathcal{P}^{s} \in  \mathbb{R}^{H \times W \times 21}
\end{equation}

\subsection{Overall architecture}

\noindent{\textbf{Integrating the priors}}. ISALux utilizes two types of priors, which are integrated into the HISA-MSA. To match the required number of channels in each HISA-MSA block, we adapt the dimensions of both the illumination and semantic maps. This adaptation is achieved using a \textit{conv}1x1 for each set of Transformer Blocks as illustrated in Figure~\ref{fig:framework}.

Since these illumination maps already have the necessary dimensions by selecting levels from the illumination pyramid, $\mathcal{P}^{i}_{k} \in \mathbb{R}^{\frac{H}{2^k} \times \frac{W}{2^k} \times {1}},\text{ with } k=\overline{0,2}$, which match the spatial resolution of the current HISA-MSA block, the convolutional layers serve only to adjust their depth without additional downsampling. 

For the semantic prior, we employ strided convolutions with a stride \(s\), where \(s \in \{1, 2, 4\}\) to downsample it and adapt its depth according to with the dimension of the correspondent HISA-MSA block.

\begin{figure*}[ht!]
        \centering
        \includegraphics[trim=0cm 0cm 0cm 0cm, clip=true, width=0.96\linewidth]{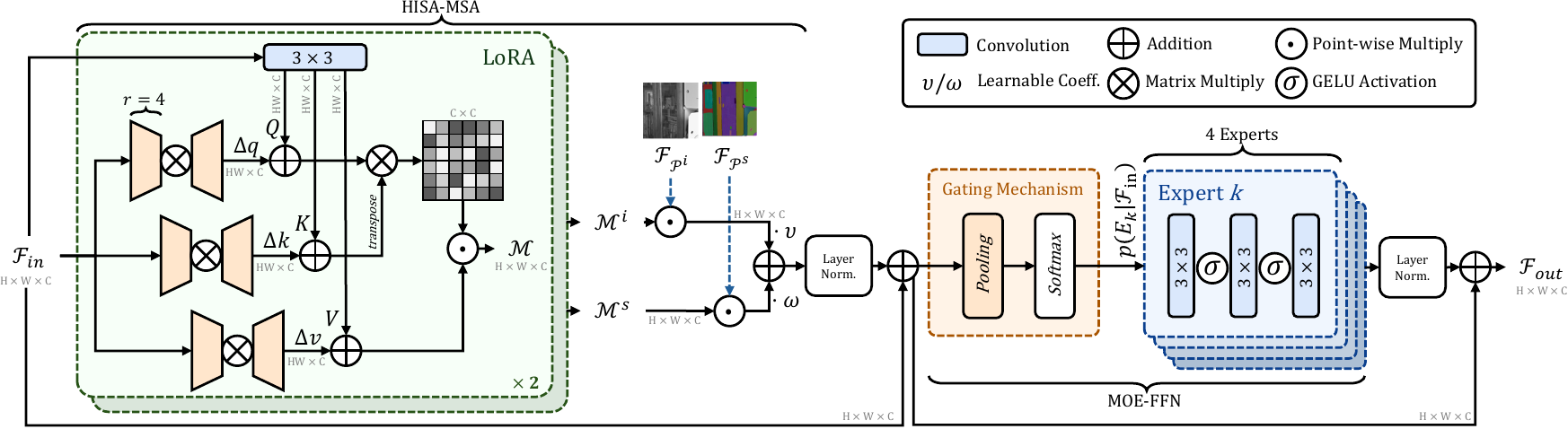}

        \caption{The architecture of HISA-MSA and the MOE-FFN blocks, our novel elements of the transformer block. The LoRA are obtained from the input feature map and added to each of the Q, K, V. The representations obtained from the self-attended modules are then passed through the top $2$ experts from MOE-FFN.}
        \label{fig:transformer}
        
    \end{figure*}

\noindent{\textbf{ISA-T}}.
The backbone of ISALux, ISA-${\mathcal{T}}$, consists of an encoder, a bottleneck, and a decoder. We denote the input as $\mathcal{I} \in \mathbb{R}^{H\times W \times3}$, the enhanced output as $\mathcal{I}_E \in \mathbb{R}^{H \times W \times 3}$, the segmentation prior as $\mathcal{P}^{s}$, and the illumination prior as $\mathcal{P}^{i}$. The enhancement process can be described as

\begin{equation}
\mathcal{I}_E = \text{ISA-}\mathcal{T}(\mathcal{I}, \mathcal{P}^s, \mathcal{P}^i)
\end{equation}

After computing the segmentation and illumination priors, the input feature \(\mathcal{I} \in \mathbb{R}^{H\times W \times3}\) is processed using a \textit{conv}3x3 with a resulting feature map, \(\mathcal{F}_{in} \in \mathbb{R}^{H\times W \times \textit{C}}\). This feature space expansion has an essential role in enriching the representation of the inputs before being processed by ISA-\(\mathcal{T}\).

\noindent Consequently, the input is passed through a U-shaped encoder-decoder series of transformer blocks for enhancement. The encoder has 2 levels, producing: $\mathcal{F}_{enc}^k \in \mathbb{R}^{\frac{H}{2^k} \times \frac{W}{2^k} \times C \cdot 2^k}, k \in \{0, 1\}$. Each output is downsampled for the next encoder. A bottleneck connects the encoder and decoder paths to enable cross-scale interaction, producing: $\mathcal{F}_{bot} \in \mathbb{R}^{\frac{H}{4} \times \frac{W}{4} \times 4C}$.

The decoder upsamples the bottleneck output and concatenates it with the encoder feature at each level. This is repeated for two levels, yielding: $\mathcal{F}_{dec}^k \in \mathbb{R}^{\frac{H}{2^k} \times \frac{W}{2^k} \times C \cdot 2^k}, k \in \{0, 1\}$. The final decoded feature $\mathcal{F}_{out} \in \mathbb{R}^{H \times W \times C}$ is passed through a \textit{conv}3x3 to obtain the enhanced image $\mathcal{I}_E \in \mathbb{R}^{H \times W \times 3}$.

\noindent{\textbf{Transformer Block}. } The core element of ISA-$\mathcal{T}$ is the Transformer Block which is used as the representation of both encoders and decoders, as seen in Figure \ref{fig:framework}.
Its structure depends on two important elements: the HISA-MSA and the MOE-FFN (Figure \ref{fig:transformer}) - the enhancement process also being aided by the intertwined residual connections and layer normalizations in the process. 

Considering \(\mathcal{F}_{in} \in \mathbb{R}^{H \times W \times C}\) as the input of the Transformer Block, \(\mathcal{F}_{out} \in \mathbb{R}^{H \times W \times C}\) as the resulting feature map, and \(k\) as the level in the U-shaped architecture, we can represent the functionality as:

\begin{gather}
\mathcal{F}_{out} = \mathcal{F}'_{in} + \text{LN}(\text{MOE\_FFN}(\mathcal{F}'_{in})) \\
\mathcal{F}'_{in} = \mathcal{F}_{in} + \text{LN}(\text{HISA\_MSA}(\mathcal{F}_{in}, \mathcal{F}_{\mathcal{P}^{i}_{k}} , \mathcal{F}_{\mathcal{P}^{s}_{k}})), \  \ \mathcal{F}_{\mathcal{P}^{i}_{k}}, \mathcal{F}_{\mathcal{P}^{s}_{k}} \in \mathbb{R}^{\frac{H}{2^k} \times \frac{W}{2^k} \times C \cdot k}, \ \ k = \overline{0,2}
\end{gather}

\noindent{\textbf{HISA-MSA}. } The use of ViT for LLIE marked a breakthrough, owing to the self-attention mechanism's role in enhancement. ISALux adopts a hybrid attention that projects illumination and segmentation-enriched attention maps, selecting their importance based on local context. To improve high-dimensional components (\( Q, K, V \)), we integrate trainable LoRA, enhancing expressiveness and regularizing the model by adding non-linearity and preventing overfitting to local patterns.

\noindent The structure of HISA-MSA is based on two parallel multi-headed self attention modules, the results of each of them being multiplied with the correspondent illumination and segmentation feature maps. The resulting enriched feature spaces are then element-wise added while each of them has a learnable parameter, namely $\alpha, \beta$.

We consider  $\mathcal{F}_{in} \in \mathbb{R}^{H \times W \times C}$ as the input, and we use a \textit{conv}3x3 to project the source for the attention elements. Then, the resulting map $\text{Proj}(\mathcal{F}_{in}) \in \mathbb{R}^{H \times W \times 3C}$ is split into 3 across the channel dimension, obtaining $Q$, $K$, $V$:

\noindent Let \(\mathcal{F}''_{\text{in}} \in \mathbb{R}^{HW \times C}\) be the reshaped \(\mathcal{F}_{\text{in}}\), and let \( \Delta q, \Delta k, \Delta v \) denote the low-rank adaptations given by, with \(\alpha_q, \alpha_k, \alpha_v \in \mathbb{R}^{C \times \frac{C}{r}}\) and \(\beta_q, \beta_k, \beta_v \in \mathbb{R}^{\frac{C}{r} \times C}\):

\begin{align*}
\small
\Delta q = (\mathcal{F}''_{\text{in}} \cdot \alpha_q)\! \cdot \beta_q; \quad
\Delta k = (\mathcal{F}''_{\text{in}} \cdot \alpha_k)\! \cdot \beta_k; \quad
\Delta v = (\mathcal{F}''_{\text{in}} \cdot \alpha_v)\! \cdot \beta_v; \quad
\Delta q, \Delta k, \Delta v \in \mathbb{R}^{HW \times C}
\end{align*}

\noindent After the adaptations are computed, each of the projections \( Q, K, V \) is updated with its corresponding adaptation. The adapted projections are then split into \( k \) heads to reduce the computational complexity of HISA-MSA:

\begin{equation}
Q, K, V = \left\{ (Q + \Delta q)_{i}, (K + \Delta k)_{i}, (V + \Delta v)_{i} \right\}_{i=1}^\mathsf{k}
\end{equation}

\noindent The \( Q \) and \( K \) components are utilized to compute the attention scores. A learnable temperature, \( \mathcal{T} \in \mathbb{R} \), is introduced to balance the relative influence of the terms and effectively capture the local context. The final attention map, \( \mathcal{M} \), is then computed by applying the softmax function to the normalized attention scores and multiplying the result with the \( V \) component:

\begin{equation}
\mathcal{M} = \text{softmax}\left(\frac{Q \cdot K^\top}{\mathcal{T}}\right) \cdot V, \quad \mathcal{M} \in \mathbb{R}^{H \times W \times C}
\end{equation}

\noindent In each Transformer Block, we use two parallel self-attention modules, producing outputs $\mathcal{M}^i$ and $\mathcal{M}^s$. These are enhanced by element-wise multiplication with illumination and semantic maps $\mathcal{F}_{\mathcal{P}^{i}}, \mathcal{F}_{\mathcal{P}^{s}} \in \mathbb{R}^{H \times W \times C}$. The resulting enriched maps are then combined using learnable weights $\boldsymbol{\upsilon}, \boldsymbol{\omega} \in \mathbb{R}$ to produce the final attention map $\mathcal{M}_\mathcal{E}$.

\begin{equation}
\mathcal{M}_\mathcal{E} = \boldsymbol{\upsilon} \cdot (\mathcal{M}^i \odot \mathcal{F}_{\mathcal{P}^{i}}) + \boldsymbol{\omega} \cdot (\mathcal{M}^s \odot \mathcal{F}_{\mathcal{P}^{s}}), \quad \mathcal{M}_\mathcal{E} \in \mathbb{R}^{H \times W \times C}
\end{equation}


\noindent{\textbf{MOE-FFN}.} Currently, the MoE~\cite{mixture_of_experts_1991, mixture_of_experts_2017} is widely used in LLMs for its efficiency and effectiveness. Its conditional activation mechanism enables selective expert use based on input context, optimizing performance and resource usage. In our work, we leverage MoE to enhance the FFN. ISALux employs a sparse MoE-FFN, which processes features identified as relevant in both local and global contexts by HISA-MSA. MoE’s ability to assign experts to different tasks is ideal for LLIE, as each expert learns deficiencies caused by insufficient lighting.

The main components of the MOE-FFN are the gating mechanism, which calculates the probability of each expert, and the experts themselves, which are represented by the following sequence of layers: \textit{conv}1x1, \textit{conv}3x3, \textit{conv}1x1 each of the previous being activated by the Gaussian Error Linear Unit (GELU) being denoted as $\sigma$, where $\mathcal{E}_i(\mathcal{F}_{in}) \in \mathbb{R}^{H \times W \times C}$.

\noindent
\hspace{0pt}
\begin{equation}
\mathcal{E}_i(\mathcal{F}_{in}) = \textit{conv}\text{1x1} \left( \sigma \left( \textit{conv}\text{3x3} \left( \sigma \left( \textit{conv}\text{1x1} \left( \mathcal{F}_{in} \right) \right) \right) \right) \right)
\end{equation}

\noindent Therefore, considering $\mathcal{F}'_{\text{in}} \in \mathbb{R}^{H \times C \times W}$ as the feature map resulting after the input has been enhanced by HISA-MSA, and $N$ as the number of experts, we can formulate the functionality of the gating mechanism as:

\begin{equation}
\mathcal{G} = \text{softmax}(\mathcal{W}_g \cdot \textit{avg\_pool}(\mathcal{F}'_{\text{in}})), \quad \mathcal{G} \in \mathbb{R}^{N},
\end{equation}

\noindent where $\mathcal{W}_g \in \mathbb{R}^{C \times N}$ represents the gating weights, and $\mathcal{G}$ represents the gating scores for $N$ experts.

\begin{table*}[ht]
\centering
\scriptsize
\renewcommand{\arraystretch}{0.95}
\setlength{\tabcolsep}{6pt}
\resizebox{\textwidth}{!}{
\begin{tabular}{l|c|cc|cc|cc|cc|cc|cc}
\toprule
\textbf{Methods} & \textbf{Params} & \multicolumn{2}{c|}{\textbf{LOL-v1}} & \multicolumn{2}{c|}{\textbf{LOL-v2-R}} & \multicolumn{2}{c|}{\textbf{LOL-v2-S}} & \multicolumn{2}{c|}{\textbf{SDSD-in}} & \multicolumn{2}{c|}{\textbf{SDSD-out}} & \multicolumn{2}{c}{\textbf{Average}} \\ 
\midrule
 & (M) & \textbf{PSNR} & \textbf{SSIM} & \textbf{PSNR} & \textbf{SSIM} & \textbf{PSNR} & \textbf{SSIM} & \textbf{PSNR} & \textbf{SSIM} & \textbf{PSNR} & \textbf{SSIM} & \textbf{PSNR} $\uparrow$ & \textbf{SSIM} $\uparrow$ \\ 
\midrule
EnGAN \scriptsize{TIP '21} \cite{KinD}      & 114.35  & 20.00 & 0.691 & 18.23 & 0.617 & 16.57  & 0.734 & 20.02 & 0.604 & 20.10 & 0.616 & 18.98 & 0.652 \\
Restormer \scriptsize{CVPR '22} \cite{Restormer} & 26.1 & 22.43 & 0.823 & 19.94 & 0.827 & 21.41 & 0.830 & 25.67 & 0.827 & 24.79 & 0.802 & 22.85 & 0.822 \\
MAXIM \scriptsize{CVPR '22} \cite{KinD}      & 14.1  & 23.43 & 0.863 & - & - & - & - & - & - & - & - & 23.43 & 0.863 \\
SNR-Net \scriptsize{CVPR '22}     & 4.01  & 24.61 & 0.842 & 21.48 & 0.849 & 24.14 & 0.928 & 29.44 & 0.894 & 28.66 & 0.866 & 25.87 & 0.876 \\
LLFlow \scriptsize{AAAI '22} \cite{LLFlow} & 37.7  & 25.13 & 0.872 & 26.20 & 0.888 & 24.80 & 0.919 & - & - & - & - & 25.38 & 0.893 \\
LLFormer \scriptsize{AAAI '23} \cite{LLFormer} & 24.6 & 25.75 & 0.823 & 26.19 & 0.819 & 28.00 & 0.927 & - & - & - & - & 26.64 & 0.856 \\
DiffLight \scriptsize{CVPR '24} \cite{Restormer} & 51.7 & 25.85 & 0.876  & 26.19 & 0.819 & 28.00 & 0.927 & - & - & - & - & 26.68 & 0.874 \\
ExpoMamba \scriptsize{ICML '24} \cite{Restormer} & 41 & 25.77 & 0.860 & 28.04 & 0.885  & - & - & - & - & - & - & 26.91 & 0.873 \\
LL-SKF \scriptsize{CVPR '23} \cite{Restormer} & 39.91 & 26.79 & 0.879 & 28.45 & \underline{0.905} & 29.11 & 0.953 & - & - & - & - & 28.12 & \textbf{0.912} \\
Retinexformer \scriptsize{ICCV '23} \cite{Restormer} & 1.61 & 27.18 & 0.850 & 27.71 & 0.856 & 29.04 & 0.939 & 29.77 & \underline{0.896} & 29.84 & 0.877 & 28.71 & 0.884 \\
GLARE \scriptsize{ECCV '24} \cite{GLARE} & 44.04 & \underline{27.35} & \textbf{0.883} & \underline{28.98} & \underline{0.905} & \underline{29.84} & \textbf{0.958} & \underline{30.10} & \underline{0.896} & \underline{30.85} & \underline{0.884} & \underline{29.42} & 0.905 \\
\midrule
\textbf{ISALux}       & 1.85  & \textbf{27.63} & \underline{0.881} & \textbf{29.76} & \textbf{0.908} & \textbf{30.78} & \underline{0.956} & \textbf{30.67} & \textbf{0.909} & \textbf{31.58} & \textbf{0.895}  & \textbf{30.08} & \underline{0.910} \\
\bottomrule
\end{tabular}
}

\caption{Quantitative results on the LOL and SDSD datasets. For better comparison, we also include the average PSNR and SSIM. The methods are ordered in ascending order based on the PSNR result on LOL-v1. Best results are in \textbf{bold}, second best are \underline{underlined}.}
\label{tab:quantitative_results}

\end{table*}

After computing $\mathcal{G} \in \mathbb{R}^{N}$, we perform a Top-k expert selection based on the probabilities $\mathcal{G} = \{p(\mathcal{E}_k \mid \mathcal{F}'_{\text{in}}) \mid k \in \{1, \dots, N\}\}$. For ISALux, two experts proved optimal, as adding a third increased computational cost with negligible PSNR improvement ($\approx$0.05~dB). The top-2 experts are selected as follows:

\begin{equation}
   \mathcal{E}_k = \text{Top}_k(\mathcal{G}, k), \ k=2; \ \ \text{with} \ \ \mathcal{F}_{\text{out}} = \sum_{i=1}^k p(\mathcal{E}_i \mid \mathcal{F}'_{\text{in}}) \mathcal{E}_i(\mathcal{F}'_{\text{in}}), \ \ \mathcal{F}_{\text{out}} \in \mathbb{R}^{H \times W \times C} 
\end{equation}

\subsection{Loss functions}

To aid the restoration process through context-based enhancement, we employ a hybrid loss function. Most previous LLIE methods have not leveraged semantic information to accurately model the illumination component within the feature space. In contrast, ISALux uses a weighted sum of losses: mean squared error loss ($\mathcal{L}_{2}$), a perceptually-driven semantic loss ($\mathcal{L}_{perc}$) based on VGG-19 weights~\cite{vgg}, and the Multi-Scale Structural Similarity Index Measure loss ($\mathcal{L}_{SSIM}$).

In the context of LLIE, the $\mathcal{L}_{2}$ loss offers significant advantages due to its emphasis on minimizing the Euclidean distance between predicted and ground truth images. Its smooth gradient profile facilitates stable and efficient optimization during training. 

\begin{equation}
\mathcal{L}_{\text{2}} = \frac{1}{N} \sum_{i=1}^N \| \hat{I}_i - I_i \|_2^2
\end{equation}

\noindent The perceptual consistency loss $\mathcal{L}_\text{perc}$ leverages a pre-trained VGG-19 model to quantify deviations in high-level feature space, ensuring perceptual fidelity between predicted and reference images. Similarly, $\mathcal{L}_{SSIM}$ captures perceptual similarity by evaluating luminance, contrast, and structural information across multiple scales. Let $\Phi$ denote the VGG-19 feature extractor and $\| \cdot \|_1$ the $\ell_1$ norm:

\begin{equation}
    \small
    \mathcal{L}_\text{perc}(I, \hat{I}) = \frac{1}{M} \sum_{i,j} \left\| \Phi(I(i,j)) - \Phi(\hat{I}(i,j)) \right\|_1, \quad
    \mathcal{L}_{SSIM}(I, \hat{I}) = 1 - \prod_{j=1}^M \text{SSIM}(I^{(j)}, \hat{I}^{(j)})^{\alpha_j}
\end{equation}

\noindent The MS-SSIM loss evaluates perceptual similarity across multiple scales by computing the SSIM between the ground truth \( I^{(j)} \) and predicted \( \hat{I}^{(j)} \) images at each scale. The overall loss is the product of these SSIM values, weighted by \( \alpha_j \).

%% file: sec/4_results.tex
\section{Experiments}

\begin{figure*}[ht!]
        \centering
        \includegraphics[trim=0cm 0cm 0cm 0cm, clip=true, width=0.96\linewidth]{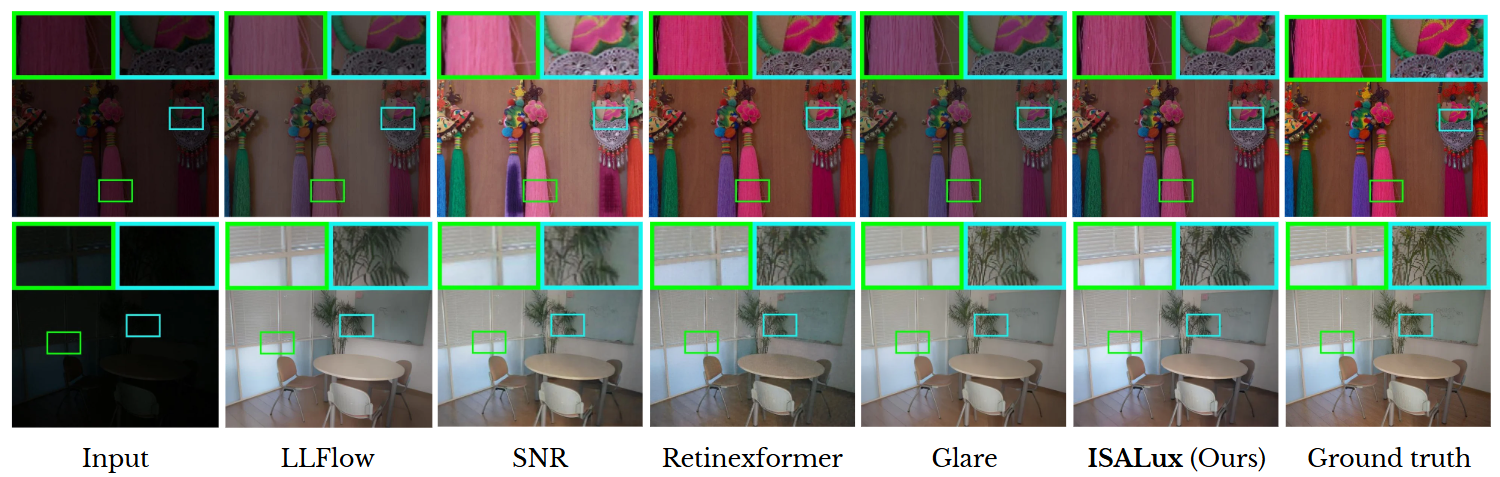}
        
        \caption{Qualitative results on the LOL dataset.}
        \label{fig:LOL_qualitative}
        
    \end{figure*}

\subsection{Datasets}


\noindent \textbf{LOL.} The LOL~\cite{RetinexNet} dataset, widely used in LLIE, contains $500$ normal/low-light image pairs at \(400 \times 600\) resolution, split into $485$ for training and $15$ for testing. Its updated version, LOLv2~\cite{Sparse}, includes LOL-v2-real ($689$/$100$ train/test split) and LOL-v2-synthetic (split of $900$/$100$).


\noindent \textbf{SDSD(in/out)}~\cite{SDSD}. We utilize the static variant of SDSD, acquired using a Canon EOS 6D Mark II camera equipped with an ND filter. The dataset is divided into indoor and outdoor subsets. Specifically, we employ low-/normal-light video pairs at $62$:$6$ and $116$:$10$ for the respective subsets.

\noindent \textbf{LOL-Blur}~\cite{LOL-Blur}. The dataset consists of $12k$ paired low-blur / normal-sharp images, which include various levels of darkness in $200$ distinct scenarios. It is split into training and testing sets in a $17$:$3$ ratio.

\noindent\textbf{LIME~\cite{LIME}, NPE~\cite{NPE}, MEF~\cite{MEF}, and DICM~\cite{DICM}.} These low-light datasets lack ground-truth references and are used for qualitative evaluation in real-world conditions.

\noindent \textbf{Implementation Details}.  
ISALux is trained for $300k$ iterations using the Adam optimizer with \(\beta_1 = 0.9\), \(\beta_2 = 0.999\). The initial learning rate of \(2 \times 10^{-4}\) increases to \(3 \times 10^{-4}\) after $92k$ iterations, then gradually decreases to \(2 \times 10^{-4}\) by $208k$, and finally to \(1 \times 10^{-6}\) by the end. Training data consists of \(256 \times 256\) patches randomly cropped from paired low-light and normal-light LOL images, with a batch size of $8$ and data augmentation via random rotation and flipping. The model minimizes a hybrid loss between outputs and ground truth. PSNR and SSIM are used on reference datasets, and (Natural Image Quality Evaluator) NIQE~\cite{NIQE} on no-reference ones.

\subsection{Quantitative Results}



\noindent \textbf{LOL \& SDSD Datasets}.
Table~\ref{tab:quantitative_results} shows ISALux achieving the best average performance with 30.08 dB / 0.910. Compared to GLARE, ISALux improves by 0.66 dB / 0.005, and over Retinexformer by 1.37 / 0.026 —while being 96\% more parameter efficient than GLARE.

\noindent \textbf{No-Reference Datasets}.
In Table~\ref{tab:combined_matched}, ISALux achieves the best NIQE average (3.34), ranking first on MEF (3.58), DICM (3.21), and NPE (3.40), and second on LIME (3.91).

\noindent \textbf{LOL Blur}.
ISALux outperforms Retinexformer 0.92 dB / 0.028, and GLARE by 0.67 dB / 0.023, confirming its robustness under both low-light and blur conditions.

\begin{table*}[h!]
\centering
\renewcommand{\arraystretch}{1.1}
\scriptsize
\setlength{\tabcolsep}{6pt}
\resizebox{\textwidth}{!}{
\begin{tabular}{l|c c c c c||l|c c}
\toprule
\multicolumn{6}{c||}{\textbf{NIQE $\downarrow$ on No-Reference Datasets}} & \multicolumn{3}{c}{\textbf{PSNR/SSIM $\uparrow$ on LOL Blur}} \\
\cmidrule{1-6} \cmidrule{7-9}
\textbf{Method} & MEF & LIME & DICM & NPE & Avg & \textbf{Method} & PSNR & SSIM \\
\midrule
SNR-Net \cite{SNR-Net} \footnotesize{CVPR '22}      & 4.14 & 5.51 & 4.62 & 4.36 & 4.63 & DeblurGAN-v2 \cite{DeblurGAN-v2} \footnotesize{ICCV '19}     & 22.30 & 0.745 \\
LLFlow \cite{LLFlow} \footnotesize{AAAI '22}         & 3.92 & 5.29 & 3.78 & 4.16 & 4.25 & Retinexformer \cite{Retinexformer} \footnotesize{ICCV '23}   & 22.90 & 0.824 \\
LL-SKF \cite{LLSKF} \footnotesize{CVPR '23}          & 4.03 & 5.15 & 3.70 & \underline{4.08} & 4.24 & LEDNet \cite{LOL-Blur} \footnotesize{ECCV '22}              & 25.74 & 0.850 \\
RFR \cite{RFR} \footnotesize{CVPR '23}               & 3.92 & \textbf{3.81} & 3.75 & 4.13 & \underline{3.92} & ASP-LED \cite{ASP-LED} \footnotesize{ICRA '24}               & 26.73 & 0.866 \\
GLARE \cite{GLARE} \footnotesize{ECCV '24}           & \underline{3.66} & 4.52 & \underline{3.61} & 4.19 & 3.99 & VQCNIR \cite{VQCNIR} \footnotesize{AAAI '24}                 & \underline{27.79} & \underline{0.875} \\
\textbf{ISALux (Ours)}                               & \textbf{3.58} & \underline{3.91} & \textbf{3.21} & \textbf{3.40} & \textbf{3.34} & \textbf{ISALux (Ours)}         & \textbf{28.01} & \textbf{0.903} \\
\bottomrule
\end{tabular}
}

\caption{Combined evaluation of no-reference NIQE scores (left) on MEF~\cite{MEF}, LIME~\cite{LIME}, DICM~\cite{DICM}, and NPE~\cite{NPE}, and full-reference PSNR/SSIM (right) on the LOL Blur dataset~\cite{LOL-Blur}. Best results are in \textbf{bold}, second best are \underline{underlined}.}
\label{tab:combined_matched}

\end{table*}

\subsection{Qualitative Results}

As shown in Figures \ref{fig:LOL_qualitative}, ISALux achieves uniform enhancement by leveraging semantic knowledge, preserving accurate colors without introducing artifacts, unlike SNR-Net. In contrast, Retinexformer and SNR-Net enhance images at the cost of color distortion, while LL-Flow and LL-SKF exhibit blurring or fail to preserve details, as seen in the zoomed qualitative results in Figure~\ref{fig:LOL_qualitative}. 
The qualitative results on no-reference datasets found in the \textbf{\textit{Supplementary Materials}} demonstrate higher color fidelity in real-world scenarios and improved illumination consistency compared to other methods.

%
%

\subsection{Inference time}

ISALux not only achieves high performance but also demonstrates substantial computational efficiency. To support this, we report average inference times on LOL test images (400×600 resolution). ISALux completes inference in just 105 ms, significantly outperforming other state-of-the-art methods such as Retinexformer (300 ms), LLFlow (378 ms), and GLARE (650 ms). This efficiency makes ISALux highly suitable for real-time or resource-constrained deployment scenarios.

%% file: sec/5_ablation.tex
\section{Ablation study}

\subsection{Effect of priors}
We conduct ablation studies to highlight the role of the illumination and segmentation priors, as well as the performance difference compared to a vanilla multi-headed self-attention module. The results in Table \ref{tab:merged_ablation} show that the illumination prior improves performance by 0.11 dB, while the segmentation prior adds 0.21 dB. Combining both priors leads to a 0.59 dB increase, with a 0.22 dB advantage over the no LoRA version.

\begin{table*}[h!]
\centering
\renewcommand{\arraystretch}{1.2}
\scriptsize
\resizebox{\textwidth}{!}{
\begin{tabular}{l|r r|r r||r r r|r r}
\toprule
\textbf{Ablation Setting} & \textbf{Illum} & \textbf{Seg} & \textbf{PSNR} & \textbf{SSIM} 
& \textbf{L2} & \textbf{VGG19} & \textbf{MS-SSIM} & \textbf{PSNR} & \textbf{SSIM} \\
\midrule
No Priors                      &        &        & 27.04 & 0.870 & \checkmark &           &           & 26.95 & 0.850 \\
Illum Only                     & \checkmark &        & 27.15 & 0.878 & \checkmark &           & \checkmark & 27.08 & 0.869 \\
Segmentation Only              &        & \checkmark & 27.36 & 0.879 & \checkmark & \checkmark &           & 27.14 & 0.872 \\
Illum + Seg (no LoRA)          & \checkmark & \checkmark & 27.41 & 0.880 & \checkmark & \checkmark & \checkmark & \textbf{27.63} & \textbf{0.881} \\
Illum + Seg                    & \checkmark & \checkmark & \textbf{27.63} & \textbf{0.881} & -- & -- & -- & -- & -- \\
\bottomrule
\end{tabular}
}

\caption{Unified ablation study evaluating the impact of semantic priors (illumination and segmentation) and loss function combinations on LOL dataset. Left columns show the contribution of priors; right columns evaluate various loss setups.}
\label{tab:merged_ablation}

\end{table*}

\subsection{Effect of Loss Functions}
We perform an ablation study to examine the impact of different loss functions on model performance. We evaluate $L_2$ loss, VGG19 perceptual loss, and MS-SSIM loss, both individually and in combination. The results in Table \ref{tab:merged_ablation} show that combining $L_2$ loss with either VGG19 perceptual loss or MS-SSIM provides a 0.13 dB improvement over plain $L_2$ loss. Adding MS-SSIM to both $L_2$ and VGG19 perceptual loss results in an additional 0.49 dB improvement.

%% file: sec/6_conclusion.tex
\section{Conclusion and Future Work}


This paper introduces ISALux, a novel ViT-based architecture that embeds illumination and semantic priors within the attention mechanism. By integrating low-rank matrices after computing Q, K, and V, the model efficiently captures spatial dependencies while minimizing computational costs. Additionally, a MoE-enhanced FFN dynamically adapts feature representations to effectively handle deficiencies caused by low-light conditions. Extensive experiments confirm that ISALux achieves state-of-the-art performance across multiple LLIE benchmarks, surpassing existing methods.

Since LLIE involves addressing various image deficiencies, future work will focus on identifying and mitigating noise patterns and artifacts caused by low-light conditions. This can be achieved by refining the current architecture or introducing a dedicated module for noise reduction. 